\documentclass{article}

\PassOptionsToPackage{numbers, compress, sort&compress}{natbib}

\usepackage[preprint]{neurips_2026}

\usepackage[utf8]{inputenc} 
\usepackage[T1]{fontenc}    
\usepackage{hyperref}       
\usepackage{url}            
\usepackage{booktabs}       
\usepackage{amsmath}        
\usepackage{amsfonts}       
\usepackage{nicefrac}       
\usepackage{microtype}      
\usepackage{xcolor}         
\usepackage{graphicx}
\usepackage{pifont}
\usepackage{wrapfig}
\usepackage{multirow}
\usepackage{makecell}
\usepackage{colortbl}
\usepackage{arydshln}
\usepackage{float}
\title{BOLT: Online Lightweight Adaptation for Preparation-Free Heterogeneous Cooperative Perception\thanks{Preprint. Work under review.}}

\author{%
  Kang Yang \\
  Renmin University of China \\
  \texttt{y1127238112@gmail.com} \\
  \And
  Tianci Bu \\
  National University of Defense Technology \\
  \texttt{btc010001@gmail.com} \\
  \And
  Peng Wang \\
  Renmin University of China \\
  \texttt{peng.wang@ruc.edu.cn} \\
  \AND
  Deying Li \\
  Renmin University of China \\
  \texttt{deyingli@ruc.edu.cn} \\
  \And
  Yongcai Wang\thanks{Corresponding author.} \\
  Renmin University of China \\
  \texttt{ycw@ruc.edu.cn} \\
}

\begin{document}

\maketitle

\begin{abstract}
Most existing heterogeneous cooperative perception methods depend on prior preparation like offline joint training or tailored collaborator-model adaptation. Such preprocessing is, however, generally impractical in real scenarios, as agents are usually independently trained by different developers and meet occasionally online. 
This work investigates \emph{preparation-free heterogeneous cooperative perception}, where agents use independently trained single-agent detectors without any pre-deployment coordination. We find direct cross-agent fusion under this setting greatly underperforms ego-only perception.
We present BOLT, a lightweight plug-and-play module that adapts neighboring features online via ego-as-teacher distillation, requiring only ego predictions without ground-truth labels. BOLT leverages high-confidence ego perception features to guide cross-agent feature-domain alignment, while enabling neighbors to contribute features in the ego's low-confidence regions.
With only 0.9M trainable parameters, BOLT improves AP@50 by up to 32.3 points over vanilla unadapted fusion in the preparation-free setting. It consistently outperforms ego-only results on DAIR-V2X and OPV2V, across different encoder pairs and fusion strategies. Code: \url{https://github.com/sidiangongyuan/BOLT}.

\end{abstract}

\section{Introduction}\label{sec:intro}

\begin{figure}[ht]
    \centering
    \includegraphics[width=\textwidth]{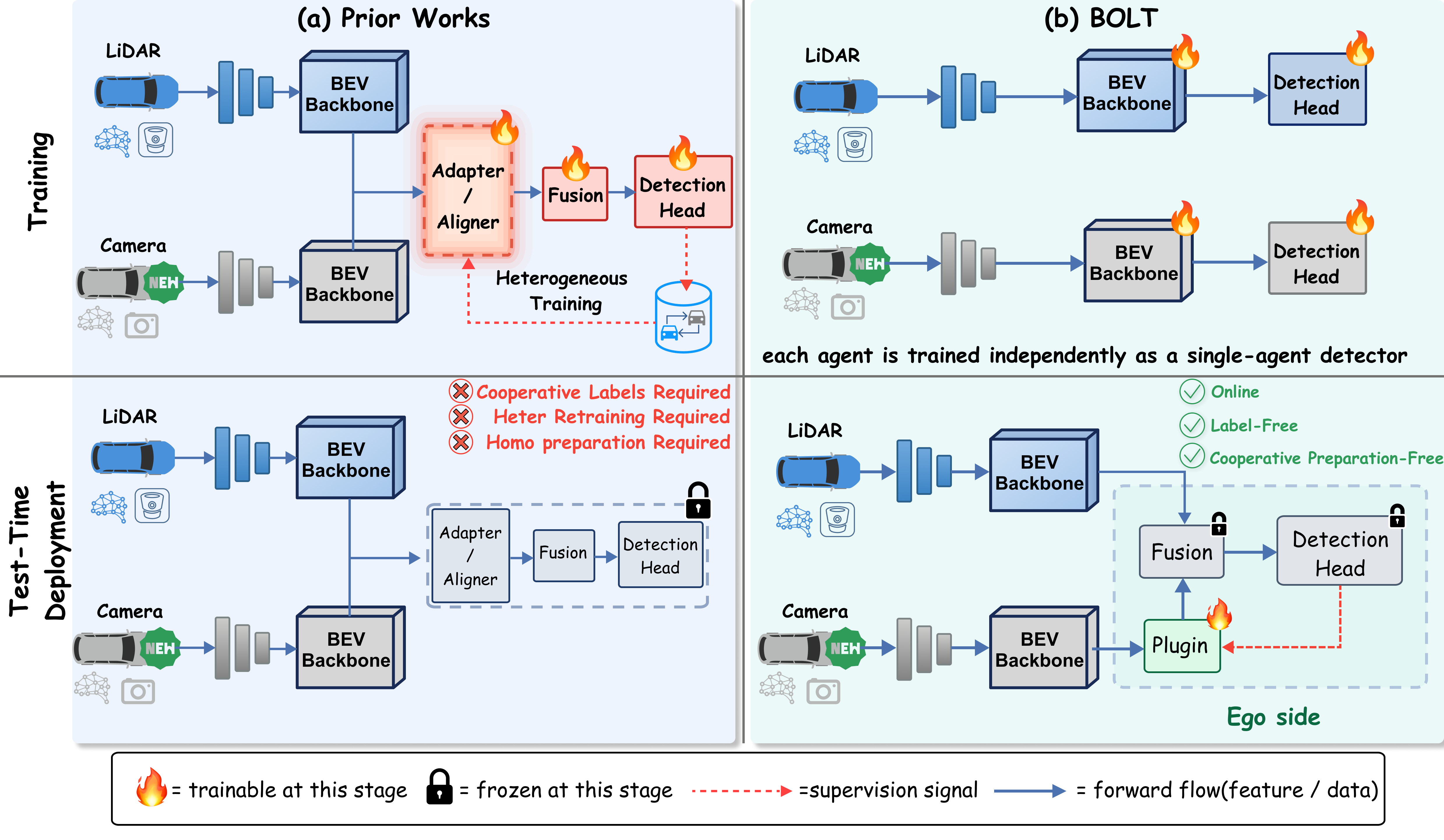}
    \caption{\textbf{Preparation-free heterogeneous cooperative perception.} \textbf{(a)~Prior works}; \textbf{(b)~BOLT}. 
    }
    \label{fig:teaser}
\end{figure}

Cooperative perception enhances autonomous driving by enabling multiple agents to share information invisible from a single agent~\cite{OPV2V,dair-v2x,collaborative-survey2,v2v4real,cooperative-development1}.
Since on-road agents are highly diverse and heterogeneous, heterogeneous cooperative perception is crucial. Currently, most existing heterogeneous cooperative perception methods follow a prepare-first paradigm. Before cooperating with target partners, they demand extra pre-deployment preparation, including offline joint training, protocol alignment, or customized adaptation data for each collaborator.
Representative methods such as HEAL~\cite{HEAL}, STAMP~\cite{stamp}, HM-ViT~\cite{HM-VIT}, PolyInter~\cite{PolyInter}, and NegoCollab~\cite{congzhangnegocollab} all rely on offline cooperative preprocessing. PHCP~\cite{PHCP} moves part of the alignment to inference time, but still builds on a cooperatively pre-trained base and warms up its plugin on an unlabeled split of each scene before reporting cooperative performance on the rest.
Such a pipeline works well for controlled experiments but fails to meet real-world deployment demands, where instant cooperation is needed. 
In practical scenarios, individual agents are typically equipped with proprietary, independently trained detectors with distinct sensors and encoders, leaving no room for joint retraining prior to cooperation.


\begin{wrapfigure}{r}{0.42\textwidth}
  \vspace{-0.6\baselineskip}
  \centering
  \includegraphics[width=0.40\textwidth]{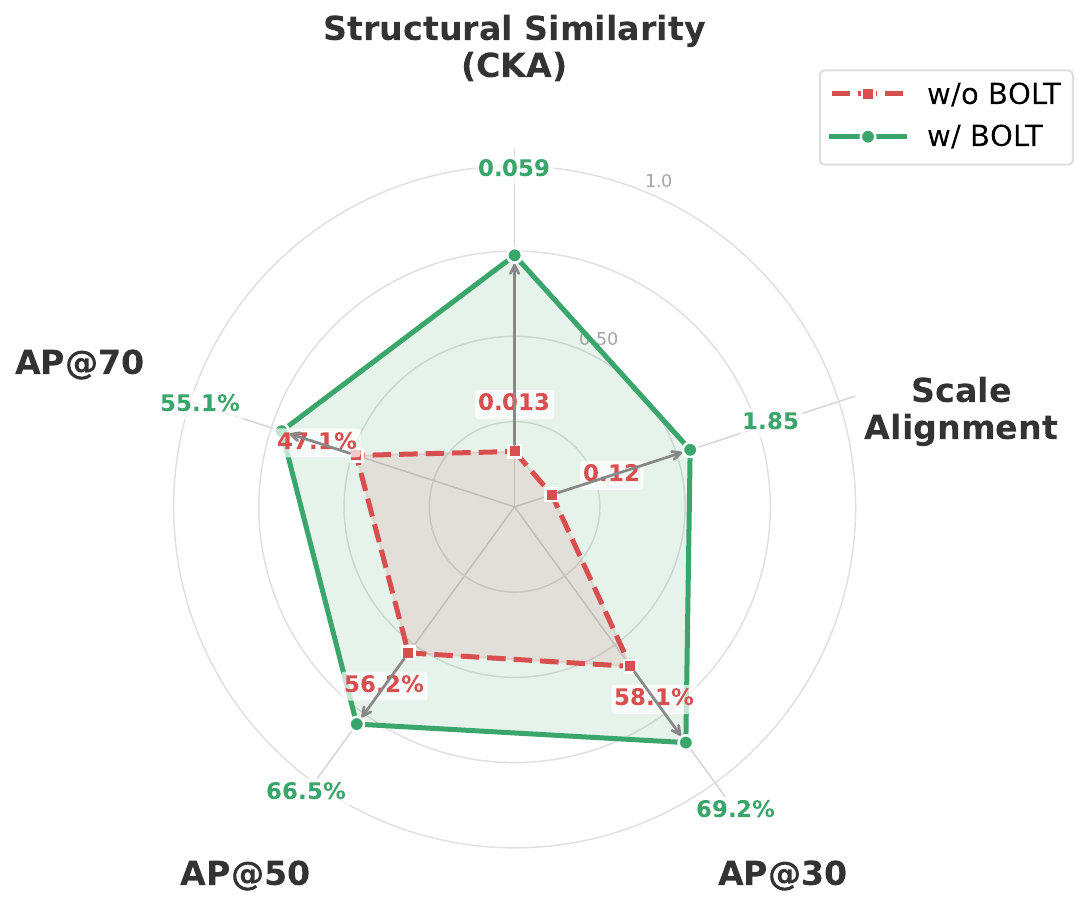}
  \caption{Effect of BOLT on DAIR-V2X with a LiDAR ego (PointPillars, abbreviated PP) and a camera neighbor (LSS-EfficientNet, abbreviated LSS-E), denoted PP$\to$LSS-E (ego$\to$neighbor) throughout. BOLT improves both feature compatibility (CKA, scale alignment) and detection accuracy (AP@30/50/70).}\label{fig:radar}
  \vspace{-0.6\baselineskip}
\end{wrapfigure}

We formalize this practical setting as preparation-free heterogeneous cooperative perception.
Unlike prior methods requiring collaborative pre-training, heterogeneity-aware retraining, or partner-specific warm-up, each agent directly adopts an independently trained single-agent detector with no pre-deployment preparation. As a result, each agent is only optimized for its own ego features and can hardly properly interpret unseen neighbor's features.
Due to severe distribution gaps between ego and heterogeneous neighbor features, naive unadapted fusion often degrades performance and even underperforms ego-only detection, as validated in Table~\ref{tab:main}.

To address this gap, this paper proposes \textbf{BOLT}. BOLT inserts a lightweight ego-side \textbf{plugin} to calibrate incoming neighbor features before fusion. Only the plugin is updated online via ego-as-teacher distillation, with encoders, fusion modules and detection heads fully frozen. This enables label-free, instant adaptation without extra cooperative constraints. 

Figure~\ref{fig:teaser} visually compares our practical deployment setting with the conventional prepare-first workflow. In \emph{Training}, prepare-first methods couple ego and neighbor agents via extra heterogeneous training to build cooperative adapters using paired collaborative data. By contrast, BOLT only trains standalone single-agent detectors. In \emph{Deployment}, prior works rely on pre-trained cooperative pipelines with strong preconditions: collaborative labels, heterogeneous retraining, and task-specific cooperative priors. BOLT instead places an adaptive plugin on the ego side: each received neighbor feature is processed by this plugin before entering the frozen fusion module, and only the plugin is updated at test time.
The design meets the deployment constraints that the original single-agent detectors remain untouched, no cooperative training data is required, and adaptation must happen online when a new collaborator appears.


To implement BOLT, our key insight is that the ego detector itself can act as a feature alignment teacher for neighboring agents.
The agent and the neighbors generally have large overlapped regions. 
In regions where the ego achieves high detection confidence, the overlapped neighbor feature representations are aligned to the ego's feature domain, with reliable ego features serving as the alignment target.
For ego-uncertain regions, neighbors supplement their perceptual information following the alignment transformation learned from high-confidence ego areas.
This design yields a lightweight plugin module deployed prior to feature fusion, enabling adaptive cross-agent \emph{feature alignment} and \emph{evidence completion}.
Updated online via ego-as-teacher distillation, it needs no shared labels or pre-deployment collaborative training.
As shown in Figure~\ref{fig:radar}, this framework improves both feature compatibility and detection accuracy.
We define the performance decline from ego-only baseline to raw unadapted cooperation as the \emph{single-to-cooperative transition gap}.

Our contributions are threefold:
\textbf{(1)}~We identify and formalize preparation-free heterogeneous cooperative perception, a deployment-oriented setting in which independently trained agents must cooperate without any pre-deployment cooperative preparation.
\textbf{(2)}~We propose BOLT, an online label-free adaptation method that optimizes a plugin via ego-as-teacher distillation, and adds this small ego-side plugin to process each neighbor feature before fusion.
\textbf{(3)}~We show across two benchmarks, multiple combinations of agent backbones, and multiple fusion strategies that BOLT consistently turns degraded preparation-free cooperation into useful cooperation, yielding up to $+32.3$ AP@50 over unadapted fusion while consistently surpassing the ego-only baseline.

\section{Related Work}

\paragraph{Cooperation under shared-protocol assumptions.}
A single agent cannot observe what its viewpoint occludes, and cooperative perception addresses this limitation by sharing information across agents~\cite{OPV2V,dair-v2x,collaborative-survey2,v2v4real,cooperative-development1,v2xreal}.
Existing methods differ mainly in where fusion happens: early fusion shares raw sensor data~\cite{early1,early2}, late fusion merges detection outputs~\cite{late1,late2}, and intermediate fusion exchanges learned features~\cite{Fcooper,OPV2V,v2vnet,where2comm,CoBEVT,how2comm}.
Intermediate fusion has become the dominant choice because it balances accuracy and communication cost well.
Later work improved communication efficiency~\cite{where2comm,how2comm,CodeFilling,ACCO,EIMC,TransIFF,quest}, robustness to localization error~\cite{conformal_uncertainty, sparsealign, Coalign,ROCO}, and asynchronous communication~\cite{CoBEVFlow,SyncNet,Traf-align}.
The common assumption is that agents are jointly trained under a homogeneous protocol.

\paragraph{Bridging heterogeneous encoders before deployment.}
Independently developed agents rarely share a common sensing and representation protocol, because they use different sensor modalities, perception models, or feature spaces, and cooperative inference must somehow reconcile these mismatched representations.
Existing methods mainly make such agents compatible in two ways.
One line of work builds a shared protocol or common representation before deployment, for example through unified heterogeneous fusion modules~\cite{HM-VIT}, shared feature spaces with backward alignment~\cite{HEAL}, adapter-reverter mappings to a protocol domain~\cite{stamp}, learned interpreters and negotiated common representations~\cite{PolyInter,congzhangnegocollab}, or codebook-based cross-modal translation for modality isolation~\cite{CodeAlign}.
Another line moves part of the alignment to inference time.
PHCP~\cite{PHCP} belongs to this second category: it performs label-free inference-time adaptation, while formulating the setting as few-shot unsupervised adaptation with a small amount of unlabeled collaborator data before collaborative reporting.
Table~\ref{tab:related} summarizes these methods.
By contrast, we study a deployment-oriented preparation-free setting where cooperation starts directly from independently trained single-agent detectors, without pre-deployment cooperative preparation.
Consequently, the cooperative inference pipeline is not preconditioned on multi-agent interaction: during training it receives only single-agent inputs, and at test time each sample is predicted before being used for adaptation.

\begin{table*}[t]
\caption{Comparison of heterogeneous cooperative perception methods under deployment-relevant properties. \textbf{Preparation-Free} = a target collaborator can be deployed without any cooperative preparation (offline cooperative training, collaborator-specific support set, etc.). \textbf{Interface Handling} summarizes where or how cross-modal incompatibility is resolved.}\label{tab:related}
\centering
\footnotesize
\setlength{\tabcolsep}{3.2pt}
\renewcommand{\arraystretch}{1.05}
\begin{tabular*}{\textwidth}{@{\extracolsep{\fill}}lcc>{\centering\arraybackslash}p{4.4cm}cc}
\toprule
\textbf{Method} & \textbf{Venue} & \makecell[c]{\textbf{Preparation-}\\\textbf{Free}} & \makecell[c]{\textbf{Interface}\\\textbf{Handling}} & \makecell[c]{\textbf{Label-}\\\textbf{Free}} & \textbf{Online} \\
\midrule
HM-ViT & ICCV'23 & \ding{55} & shared heter.\ fusion & \ding{55} & \ding{55} \\
HEAL & ICLR'24 & \ding{55} & local aligner & \ding{55} & \ding{55} \\
STAMP & ICLR'25 & \ding{55} & local adapter + ego reverter & \ding{55} & \ding{55} \\
PolyInter & CVPR'25 & \ding{55} & ego interpreter & \ding{55} & \ding{55} \\
PHCP & ICCV'25 & \ding{55} & ego refinement & \checkmark{} & \checkmark{} \\
NegoCollab & NeurIPS'25 & \ding{55} & sender-receiver translators & \ding{55} & \ding{55} \\
CodeAlign & arXiv'26 & \ding{55} & codebook translation & \ding{55} & \ding{55} \\
\midrule
\textbf{BOLT (Ours)} & -- & \checkmark{} & ego plugin & \checkmark{} & \checkmark{} \\
\bottomrule
\end{tabular*}
\end{table*}

\paragraph{Online adaptation at test time.}
When a deployed model encounters inputs that do not match its training distribution, offline retraining is often impossible, and adaptation must happen online from unlabeled test inputs.
Test-time training (TTT) responds to this need by updating the model at inference time.
Early TTT work combines the main task with self-supervised objectives such as rotation prediction~\cite{TTT}, while fully test-time adaptation methods later showed that adaptation can be driven directly by test-time objectives such as entropy minimization~\cite{Tent}.
Subsequent studies further analyze when self-supervised test-time updates help or fail~\cite{TTTpp}, and recent extensions broaden the paradigm beyond visual recognition to settings such as large language models~\cite{TLM} and out-of-distribution recommender systems~\cite{DT3OR}.
BOLT follows the same high-level idea of online adaptation, but targets a different deployment challenge: enabling cooperation when independently trained heterogeneous encoders must interact through a frozen fusion module.
For adaptation, BOLT updates the plugin using the ego detector's predictions as teacher signals for cooperative detection, without labels, source-data retrieval, or additional cooperative pretraining.

\section{Method}

\begin{figure}[htbp]
    \centering
    \includegraphics[width=\textwidth]{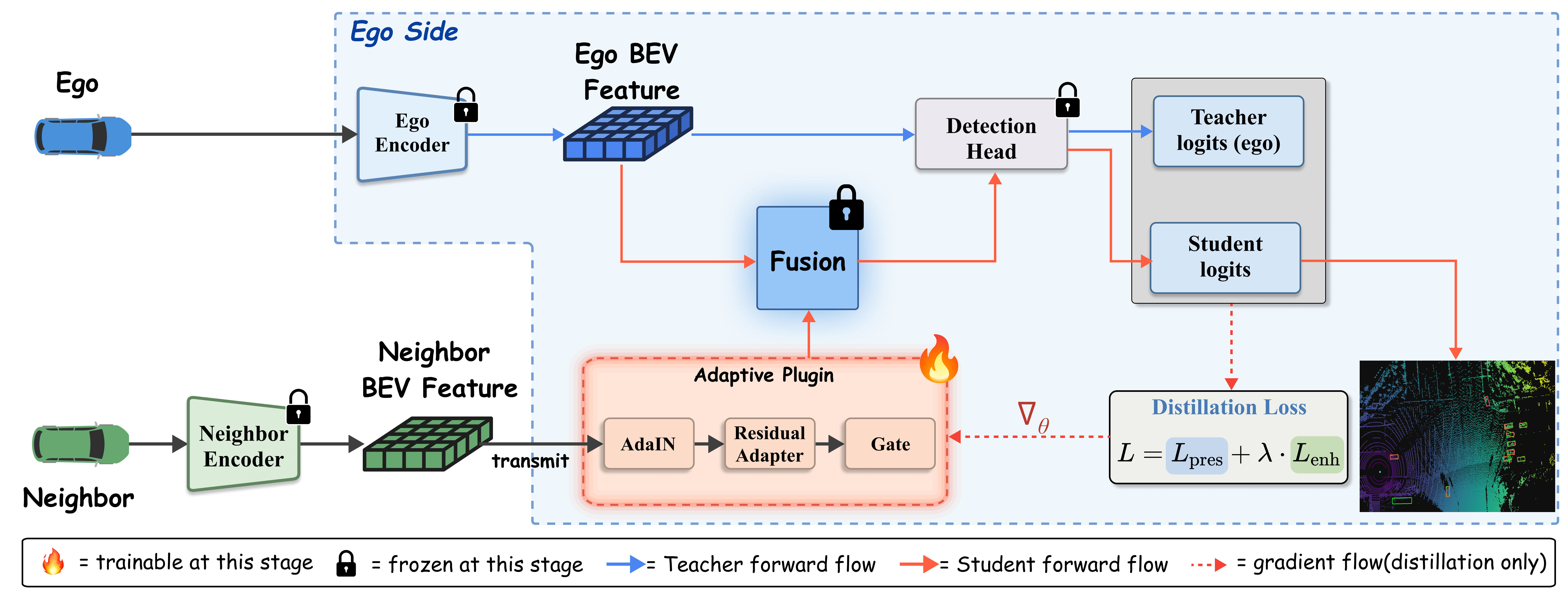}
    \caption{Framework overview. The ego agent's frozen single-agent path (top) produces teacher predictions. The neighbor agent transmits its BEV feature to the ego side, where the received feature passes through the adaptive plugin $\mathcal{P}_{\boldsymbol{\theta}}$ (AdaIN $\to$ residual adapter $\to$ per-channel gate) before entering the frozen fusion module. Only $\mathcal{P}_{\boldsymbol{\theta}}$ is updated online via ego-as-teacher distillation; all other components remain frozen.}\label{fig:framework}
\end{figure}

BOLT is built around three requirements imposed by the preparation-free setting: preserve the ego agent's working single-agent path, adapt only the ego-side stream that processes received neighbor features, and learn without labels or offline cooperative retraining.
Accordingly, the original encoders, fusion module, and detection head remain frozen, and all adaptation capacity is concentrated in a small online-updated plugin.
Figure~\ref{fig:framework} summarizes the overall design.

\subsection{Problem Formulation}

Consider $N$ agents, each equipped with an independently pre-trained encoder $\mathcal{E}_i$ that maps raw sensor data $\mathbf{X}_i$ to a bird's-eye-view (BEV) feature map $\mathbf{F}_i = \mathcal{E}_i(\mathbf{X}_i) \in \mathbb{R}^{C \times H \times W}$.
A designated \emph{ego} agent $e$ (the agent that performs fusion and inference) aggregates features from its collaborators through a fusion module $\mathcal{G}$ and produces detections via a detection head $\mathcal{H}$.
Cooperation requires that each non-ego agent $i$ first transmits its feature to the ego agent; we denote the feature received on the ego side by $\mathbf{F}_{i\to e}$, to make explicit that fusion happens after this transmission step.
In the standard cooperative perception pipeline, the detection output is
\begin{equation}\label{eq:coop}
  \hat{\mathbf{Y}}_\mathrm{coop} = \mathcal{H}\!\bigl(\mathcal{G}(\mathbf{F}_e, \, \{\mathbf{F}_{i\to e}\}_{i \neq e})\bigr),
\end{equation}
where the fusion happens entirely on the ego side and $\mathcal{G}$ is jointly trained on multi-agent cooperative data, learning to fuse heterogeneous features into a shared representation.

We study the preparation-free setting, defined by two constraints:
\textbf{(i)}~each encoder $\mathcal{E}_i$ may use a different architecture or sensor modality, and is trained independently as part of a standalone single-agent detector;
\textbf{(ii)}~cooperation is attempted at deployment through a frozen ego-side cooperative inference pipeline, whose fusion module $\mathcal{G}$ and detection head $\mathcal{H}$ are reused without pre-deployment cooperative preparation for the current heterogeneous multi-agent interface.
In this setting, the ego agent still has a reliable standalone prediction,
\begin{equation}\label{eq:ego}
  \hat{\mathbf{Y}}_\mathrm{ego} = \mathcal{D}_e(\mathbf{X}_e),
\end{equation}
where $\mathcal{D}_e$ denotes the ego agent's pre-trained single-agent detector, including prediction decoding and post-processing.
However, directly applying Eq.~\ref{eq:coop} without adaptation in this preparation-free setting often degrades cooperative detection below the ego-only baseline, because the frozen cooperative inference pipeline was never prepared for the current heterogeneous multi-agent interface and therefore misinterprets the incoming neighbor features rather than fusing them productively.

To close this gap, BOLT considers one ego feature $\mathbf{F}_e$ and one received collaborator feature, written $\mathbf{F}_n \!\equiv\! \mathbf{F}_{n\to e}$ for brevity (we drop the $n\to e$ subscript whenever it is unambiguous).
All encoders, the fusion module, and the detection head remain frozen.
Our method updates, online at test time, an adaptive plugin $\mathcal{P}_{\boldsymbol{\theta}}$ that adapts $\mathbf{F}_n$ into a representation more compatible with the frozen cooperative inference pipeline, yielding the adapted cooperative prediction
\begin{equation}\label{eq:objective}
  \hat{\mathbf{Y}}_\mathrm{adapt} = \mathcal{H}\!\bigl(\mathcal{G}(\mathbf{F}_e,\; \mathcal{P}_{\boldsymbol{\theta}}(\mathbf{F}_n, \mathbf{F}_e))\bigr).
\end{equation}
The objective of $\mathcal{P}_{\boldsymbol{\theta}}$ is to recover, and ideally surpass, the ego-only baseline without using any ground-truth labels or cooperative training data.

\subsection{Adaptive Plugin}

The plugin $\mathcal{P}_{\boldsymbol{\theta}}$ in Eq.~\ref{eq:objective} must turn the neighbor feature $\mathbf{F}_n$ into something the frozen fusion module $\mathcal{G}$ can productively consume, using only the ego feature's per-channel statistics $(\boldsymbol{\mu}_e, \boldsymbol{\nu}_e)$ as a lightweight reference.
Because adaptation begins from the first test sample, a large early perturbation would push the frozen pipeline off its training manifold before any useful gradient arrives \cite{moco}; we therefore require $\mathcal{P}_{\boldsymbol{\theta}}$ to act as the identity at initialization:
\begin{equation}\label{eq:identity}
  \mathcal{P}_{\boldsymbol{\theta}_0}(\mathbf{F}_n, \mathbf{F}_e) = \mathbf{F}_n,
\end{equation}
where $\boldsymbol{\theta}_0$ denotes the parameter values at initialization, so that each online update adds a small, controllable correction rather than replacing the representation, keeping optimization stable from the first gradient step.
Concretely, $\mathcal{P}_{\boldsymbol{\theta}}$ composes three stages, namely statistical alignment, semantic transformation, and selective gating, each parameterized so that a pass-through configuration is reachable and jointly recovers Eq.~\ref{eq:identity} at $\boldsymbol{\theta}_0$.

\textbf{Adaptive instance normalization.}\quad
Because heterogeneous encoders often differ first in channel-wise feature scale and offset, we begin by aligning the neighbor feature statistics to the ego reference via AdaIN~\cite{huang2017adain}.
Let $\boldsymbol{\mu}, \boldsymbol{\nu} \in \mathbb{R}^{C \times 1 \times 1}$ denote per-channel spatial mean and standard deviation.
Rather than replacing $\mathbf{F}_n$ outright, we blend the AdaIN-aligned feature with the original through a learnable per-channel coefficient:
\begin{equation}\label{eq:adain}
  \mathbf{F}_\mathrm{base} = \mathbf{F}_n + \boldsymbol{\alpha} \odot \!\left(\frac{\mathbf{F}_n - \boldsymbol{\mu}_n}{\boldsymbol{\nu}_n} \cdot \boldsymbol{\nu}_e + \boldsymbol{\mu}_e - \mathbf{F}_n\right),
\end{equation}
where $\boldsymbol{\alpha} \in [0,1]^{C}$ is a learnable per-channel blending coefficient, parameterized as $\boldsymbol{\alpha} = \sigma(\boldsymbol{\alpha}_0)$, and $\sigma$ denotes the sigmoid function.
The statistics $\boldsymbol{\mu}_e$, $\boldsymbol{\nu}_e$, $\boldsymbol{\mu}_n$, and $\boldsymbol{\nu}_n$ are computed on the fly from the current ego and neighbor BEV features over spatial dimensions, and a small $\epsilon$ is added to $\boldsymbol{\nu}_n$ for numerical stability.
This stage introduces no learnable parameters beyond $\boldsymbol{\alpha}$.
We initialize $\boldsymbol{\alpha}_0 {=} {-}10$ so that $\boldsymbol{\alpha} \approx 0$ at the start, leaving $\mathbf{F}_\mathrm{base}\!\approx\!\mathbf{F}_n$ until the AdaIN correction is gradually opened up by adaptation.

\textbf{Residual CNN adapter.}\quad
Statistical alignment alone cannot bridge the semantic gap between independently trained encoders.
We therefore add a small convolutional adapter to learn a residual correction on top of $\mathbf{F}_\mathrm{base}$:
\begin{equation}\label{eq:adapter}
  \boldsymbol{\Delta} = W_\mathrm{out}\!\bigl(\mathrm{ResBlocks}(\mathrm{ReLU}(\mathrm{GN}(W_\mathrm{in}\, \mathbf{F}_\mathrm{base})))\bigr),
\end{equation}
where $W_\mathrm{in} \in \mathbb{R}^{d \times C}$ projects to a hidden dimension $d$, $L$ residual blocks with $3{\times}3$ convolutions and group normalization capture spatial-channel interactions, and $W_\mathrm{out} \in \mathbb{R}^{C \times d}$ projects back.
Following the standard zero-init scheme used in residual adapters~\cite{houlsby2019adapter} and LoRA-style modules~\cite{hu2022lora}, we set the output projection $W_\mathrm{out}$ to zero so that $\boldsymbol{\Delta}{=}\mathbf{0}$ at initialization.
Zero-initializing $W_\mathrm{in}$ instead would also yield $\boldsymbol{\Delta}{=}\mathbf{0}$ but block the gradient through the entire adapter; placing the zero at the output keeps $W_\mathrm{in}$ and the residual blocks on a non-degenerate gradient path, so the adapter starts learning from the first step.

\textbf{Per-channel gate.}\quad
A learnable gate $\mathbf{g} \in [0,1]^{C}$, parameterized as $\mathbf{g} = \sigma(\mathbf{g}_0)$ with $\mathbf{g}_0\!=\!\mathbf{0}$ so that $\mathbf{g}\!=\!0.5$ at initialization, controls how much residual correction is injected into each channel:
\begin{equation}\label{eq:gate}
  \mathcal{P}_{\boldsymbol{\theta}}(\mathbf{F}_n, \mathbf{F}_e) = \mathbf{F}_\mathrm{base} + \mathbf{g} \odot \boldsymbol{\Delta}.
\end{equation}
The combined initialization ($\boldsymbol{\alpha} {\approx} 0$, $\boldsymbol{\Delta} {=} \mathbf{0}$, $\mathbf{g} {=} 0.5$) satisfies the identity property in Eq.~\ref{eq:identity}: when $\boldsymbol{\alpha} {=} 0$, $\mathbf{F}_\mathrm{base} {=} \mathbf{F}_n$; when $\boldsymbol{\Delta} {=} \mathbf{0}$, the gate has no effect, yielding $\mathcal{P}_{\boldsymbol{\theta}_0}(\mathbf{F}_n, \mathbf{F}_e) {=} \mathbf{F}_n$.
The plugin is agnostic to the choice of fusion module $\mathcal{G}$ and remains lightweight relative to the frozen backbone.

\subsection{Online Test-Time Training}

When the ego cooperates with $K$ collaborators, BOLT instantiates one independent plugin $\mathcal{P}_{\boldsymbol{\theta}_k}$ per collaborator $k\!\in\!\{1,\dots,K\}$, each updated online from its own received feature $\mathbf{F}_k$ paired with the shared ego feature $\mathbf{F}_e$. The plugins do not interact, so it suffices to describe the optimization for a single neighbor; we drop the index $k$ and write $\mathcal{P}_{\boldsymbol{\theta}}$ throughout the rest of this subsection.
At deployment, no labeled cooperative data is available, so $\mathcal{P}_{\boldsymbol{\theta}}$ must be adapted without ground-truth supervision; we therefore update it through \emph{online test-time training} (TTT) with ego-as-teacher distillation.

\textbf{Ego-as-teacher.}\quad
The key observation is that the ego agent already has a reliable standalone detector, so its high-confidence predictions provide a practical teacher signal on the anchors it already handles well.
This lets the plugin learn an ego-side correction to the received neighbor feature while preserving detections that the ego detector already gets right.
For each test sample $(\mathbf{X}_e, \mathbf{X}_n)$, we run the frozen ego detector and use its raw predictions as teacher signals, denoted by $(\mathbf{t}_\mathrm{cls}, \mathbf{t}_\mathrm{reg}, \mathbf{t}_\mathrm{dir})$ for classification, box regression, and heading direction, respectively.
The student output is obtained from the cooperative inference pipeline with the plugin enabled:
\begin{equation}\label{eq:student}
  (\mathbf{s}_\mathrm{cls},\, \mathbf{s}_\mathrm{reg},\, \mathbf{s}_\mathrm{dir}) = \mathcal{H}\!\bigl(\mathcal{G}(\mathbf{F}_e,\; \mathcal{P}_{\boldsymbol{\theta}}(\mathbf{F}_n, \mathbf{F}_e))\bigr).
\end{equation}
The student's prediction is saved for evaluation before the gradient update, so each sample is evaluated with a model that has never seen that sample's supervision signal.
We index anchors by $i$ in what follows, so $\mathbf{s}_\mathrm{cls}^i$ and $\mathbf{t}_\mathrm{cls}^i$ denote the student and teacher classification logits at anchor $i$, with the same convention for the regression and direction outputs.
The training objective consists of two complementary terms, separated by teacher confidence.

\textbf{Preservation loss.}\quad
This term realizes the feature-alignment principle from Sec.~\ref{sec:intro}: at anchors where the teacher is confident ($\sigma(\mathbf{t}_\mathrm{cls}^i) > \tau_\mathrm{hi}$), we match the student to the teacher so that online adaptation pulls the cooperative branch toward the ego-trained feature domain instead of damaging detections the ego detector already supports:
\begin{equation}\label{eq:preserve}
  \mathcal{L}_\mathrm{pres} = \frac{1}{|\mathcal{M}_{\tau_\mathrm{hi}}|}\sum_{i \in \mathcal{M}_{\tau_\mathrm{hi}}} \bigl[\,
    \ell_\mathrm{bce}(\mathbf{s}_\mathrm{cls}^i,\, \sigma(\mathbf{t}_\mathrm{cls}^i))
    + \ell_\mathrm{sl1}(\mathbf{s}_\mathrm{reg}^i,\, \mathbf{t}_\mathrm{reg}^i)
    + \ell_\mathrm{mse}(\mathbf{s}_\mathrm{dir}^i,\, \mathbf{t}_\mathrm{dir}^i)
  \,\bigr],
\end{equation}
where $\mathcal{M}_{\tau_\mathrm{hi}} = \{i : \sigma(\mathbf{t}_\mathrm{cls}^i) > \tau_\mathrm{hi}\}$ is the set of confident anchors and $\ell_\mathrm{bce}$, $\ell_\mathrm{sl1}$, $\ell_\mathrm{mse}$ denote binary cross-entropy, smooth-$\ell_1$, and mean squared error, respectively.

\textbf{Enhancement loss.}\quad
This term realizes the evidence-completion principle from Sec.~\ref{sec:intro} and handles anchors where the teacher is uncertain but not silent.
In these regions, we do not force the student to copy the teacher; instead, we encourage a stronger positive response when collaborator features provide complementary evidence:
\begin{equation}\label{eq:enhance}
  \mathcal{L}_\mathrm{enh} = -\frac{1}{|\mathcal{M}_\mathrm{boost}|}\sum_{i \in \mathcal{M}_\mathrm{boost}} \log\, \sigma(\mathbf{s}_\mathrm{cls}^i),
\end{equation}
where $\mathcal{M}_\mathrm{boost} = \{i : \tau_\mathrm{lo} < \sigma(\mathbf{t}_\mathrm{cls}^i) \leq \tau_\mathrm{hi}\}$ selects anchors in a moderate-confidence band.
The lower bound $\tau_\mathrm{lo}$ excludes near-zero predictions that are likely background or false positives, while the upper bound $\tau_\mathrm{hi}$ avoids anchors already handled by the preservation loss.
The remaining anchors correspond to locations where the ego detector sees a weak but non-negligible signal, for example under partial occlusion or at peripheral viewpoints.
Because the collaborator observes the scene from a different pose, its features can supply additional evidence exactly at such ambiguous locations.
This term is intentionally conservative: it acts only on anchors for which the teacher already indicates weak positive evidence.
By excluding near-zero teacher responses, limiting the band to $[\tau_\mathrm{lo}, \tau_\mathrm{hi}]$, and using a small weight $\lambda$, the enhancement term reduces the risk of introducing unsupported false positives.

\textbf{Overall optimization.}\quad
The plugin parameters $\boldsymbol{\theta}$ are updated by minimizing the total loss over each test sample:
\begin{equation}\label{eq:total}
  \boldsymbol{\theta} \leftarrow \boldsymbol{\theta} - \eta\, \nabla_{\boldsymbol{\theta}} \bigl(\mathcal{L}_\mathrm{pres} + \lambda\, \mathcal{L}_\mathrm{enh}\bigr),
\end{equation}
where $\eta$ is the learning rate and $\lambda$ controls the enhancement weight.
Only $\boldsymbol{\theta}$, comprising the AdaIN coefficient $\boldsymbol{\alpha}_0$, the adapter weights $\{W_\mathrm{in}, W_\mathrm{out}\}$ together with the residual-block parameters, and the gate $\mathbf{g}_0$, receives gradients.
The adaptation proceeds in a single online pass through the test stream, with each sample's prediction recorded before its corresponding gradient update.

\definecolor{gain}{HTML}{E8F5E9}
\definecolor{best}{HTML}{C8E6C9}
\definecolor{drop}{HTML}{FFEBEE}
\newcommand{\g}[1]{{\color{green!50!black}\scriptsize #1}}

\section{Experiments}

\subsection{Experimental Setup}

\paragraph{Setup.}
We evaluate BOLT on \textbf{DAIR-V2X}~\cite{dair-v2x} (real V2I) and \textbf{OPV2V}~\cite{OPV2V} (simulated V2V).
We consider two LiDAR encoders, PointPillars (\textbf{PP})~\cite{PointPillars} and \textbf{SECOND}~\cite{SECOND}, and two Lift-Splat-Shoot~\cite{LSS} camera encoders with EfficientNet~\cite{efficientnet} (\textbf{LSS-E}) and ResNet-50~\cite{ResNet} (\textbf{LSS-R50}) backbones; for each pair the ego encoder is fixed and the collaborator uses a different architecture.
All branches are independently trained as single-agent detectors and then plugged into the frozen PyramidFusion stack from HEAL~\cite{HEAL} with no Stage-2 cooperative alignment.
Unless otherwise stated, main-paper tables follow a strict two-agent protocol with one ego and one active collaborator, so each comparison isolates interface mismatch between independently trained encoders.
The adaptive plugin (AdaIN, residual adapter, per-channel gate) carries about 0.9M trainable parameters and is optimized via single-pass online ego-as-teacher distillation.
Appendix~\ref{app:impl} gives the reproducibility details, including encoder construction, plugin architecture, online protocol, data splits, and AP computation.
Appendix~\ref{app:efficiency} reports deployment-time latency, online-update cost, and cost accounting.

\subsection{Main Results}

\begin{table}[t]
\caption{Preparation-free cooperative perception on DAIR-V2X and OPV2V.\@
PP and SECOND are LiDAR encoders; LSS-E and LSS-R50 are Lift-Splat camera encoders with EfficientNet and ResNet-50 backbones.
\textbf{w/o Plugin} uses independently trained single-agent branches with a frozen PyramidFusion cooperative stack and no online adaptation.
\textbf{w/ Plugin} enables BOLT under the same setting.
Each result uses one ego and one active collaborator.
Green cells surpass ego-only; $\Delta$ is the AP@50 gain from the plugin.}\label{tab:main}
\centering
\resizebox{\columnwidth}{!}{%
\begin{tabular}{cl ccc ccc r}
\toprule
\multirow{2}{*}{\textbf{Ego}} & \multirow{2}{*}{\textbf{Collaborator}}
  & \multicolumn{3}{c}{\textbf{w/o Plugin}}
  & \multicolumn{3}{c}{\textbf{w/ Plugin}}
  & \multirow{2}{*}{\textbf{$\Delta$AP@50}} \\
\cmidrule(lr){3-5} \cmidrule(lr){6-8}
& & AP@30 & AP@50 & AP@70 & AP@30 & AP@50 & AP@70 & \\
\midrule
\multicolumn{9}{c}{\cellcolor{gray!8}\textbf{DAIR-V2X}~\cite{dair-v2x}} \\
\rowcolor{white}
\color{gray}PP     & \color{gray}\textit{ego-only}  & \color{gray}63.6 & \color{gray}60.0 & \color{gray}45.1 & & & & \\
\rowcolor{white}
\color{gray}SECOND & \color{gray}\textit{ego-only}  & \color{gray}66.9 & \color{gray}60.6 & \color{gray}51.9 & & & & \\
\rowcolor{white}
\color{gray}PP     & \color{gray}\textit{homo.\ coop}~\cite{HEAL} & \color{gray}82.9 & \color{gray}78.3 & \color{gray}58.9 & & & & \\
\rowcolor{white}
\color{gray}SECOND & \color{gray}\textit{homo.\ coop}~\cite{HEAL} & \color{gray}84.6 & \color{gray}80.8 & \color{gray}67.4 & & & & \\
\cdashline{1-9}
PP     & SECOND & 45.9 & 44.2 & 34.4 & \cellcolor{gain}\textbf{72.6} & \cellcolor{gain}\textbf{67.4} & \cellcolor{gain}\textbf{53.3} & \g{+23.2} \\
PP     & LSS-E  & 58.1 & 56.2 & 47.1 & \cellcolor{gain}\textbf{69.2} & \cellcolor{gain}\textbf{66.5} & \cellcolor{gain}\textbf{55.1} & \g{+10.3} \\
PP     & LSS-R50 & 51.8 & 50.1 & 41.7 & \cellcolor{gain}\textbf{69.2} & \cellcolor{gain}\textbf{65.1} & \cellcolor{gain}\textbf{52.5} & \g{+13.9} \\
SECOND & PP     & 64.7 & 62.3 & 49.5 & \cellcolor{gain}\textbf{72.9} & \cellcolor{gain}\textbf{68.5} & \cellcolor{gain}\textbf{52.5} & \g{+6.2} \\
SECOND & LSS-E  & 63.4 & 60.9 & 48.1 & \cellcolor{gain}\textbf{69.5} & \cellcolor{gain}\textbf{66.2} & \cellcolor{gain}\textbf{53.1} & \g{+5.3} \\
SECOND & LSS-R50 & 65.7 & 62.5 & 48.6 & \cellcolor{gain}\textbf{70.8} & \cellcolor{gain}\textbf{66.9} & \cellcolor{gain}\textbf{52.6} & \g{+4.4} \\
\midrule
\multicolumn{9}{c}{\cellcolor{gray!8}\textbf{OPV2V}~\cite{OPV2V}} \\
\rowcolor{white}
\color{gray}PP     & \color{gray}\textit{ego-only}  & \color{gray}78.2 & \color{gray}76.5 & \color{gray}65.1 & & & & \\
\rowcolor{white}
\color{gray}SECOND & \color{gray}\textit{ego-only}  & \color{gray}79.0 & \color{gray}77.8 & \color{gray}70.5 & & & & \\
\rowcolor{white}
\color{gray}PP     & \color{gray}\textit{homo.\ coop}~\cite{HEAL} & \color{gray}96.3 & \color{gray}95.6 & \color{gray}89.8 & & & & \\
\rowcolor{white}
\color{gray}SECOND & \color{gray}\textit{homo.\ coop}~\cite{HEAL} & \color{gray}95.7 & \color{gray}94.5 & \color{gray}85.6 & & & & \\
\cdashline{1-9}
PP     & SECOND & 65.5 & 64.9 & 55.1 & \cellcolor{gain}\textbf{84.5} & \cellcolor{gain}\textbf{83.0} & \cellcolor{gain}\textbf{71.3} & \g{+18.1} \\
PP     & LSS-E  & 72.5 & 71.3 & 62.1 & \cellcolor{gain}\textbf{86.8} & \cellcolor{gain}\textbf{84.8} & \cellcolor{gain}\textbf{72.0} & \g{+13.5} \\
PP     & LSS-R50 & 63.8 & 62.7 & 54.3 & \cellcolor{gain}\textbf{86.8} & \cellcolor{gain}\textbf{84.9} & \cellcolor{gain}\textbf{71.8} & \g{+22.2} \\
SECOND & PP     & 47.6 & 47.5 & 45.8 & \cellcolor{gain}\textbf{82.5} & \cellcolor{gain}\textbf{79.8} & \textbf{66.5} & \g{+32.3} \\
SECOND & LSS-E  & 70.7 & 70.0 & 65.2 & \cellcolor{gain}\textbf{85.0} & \cellcolor{gain}\textbf{83.1} & \textbf{70.1} & \g{+13.1} \\
SECOND & LSS-R50 & 71.7 & 70.8 & 63.6 & \cellcolor{gain}\textbf{85.2} & \cellcolor{gain}\textbf{83.1} & \textbf{70.1} & \g{+12.4} \\
\bottomrule
\end{tabular}%
}
\end{table}

\begin{figure}[t]
  \centering
  \includegraphics[width=\textwidth]{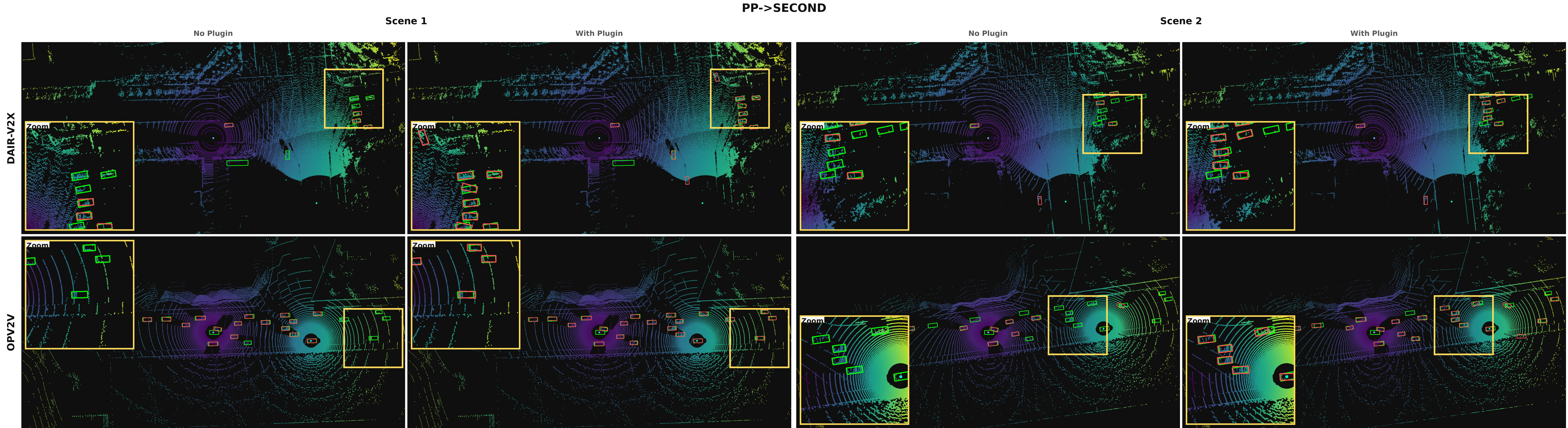}
  \caption{Qualitative BEV results. Top: DAIR-V2X; bottom: OPV2V. \textbf{Left}: w/o plugin; \textbf{right}: w/ plugin (BOLT).}\label{fig:bev}
\end{figure}

Table~\ref{tab:main} shows the same pattern on both DAIR-V2X and OPV2V: unadapted preparation-free cooperation is unreliable, but a small online adapter recovers useful cooperation.\@
Without the plugin, every evaluated pair falls below ego-only detection, confirming that the single-to-cooperative transition gap is a practical failure mode.
The degradation can be severe; on OPV2V, SECOND$\to$PP drops from 77.8 AP@50 in ego-only mode to 47.5 under unadapted preparation-free cooperation.
Figure~\ref{fig:bev} shows BEV examples aligned with Table~\ref{tab:main} on PP$\to$SECOND, where unadapted cooperation misses or mislocalizes objects that BOLT recovers; Appendix~\ref{app:qual} extends this comparison to four additional encoder pairs.

With online plugin adaptation, every evaluated pair surpasses ego-only performance, with PP$\to$SECOND reaching 67.4 AP@50 on DAIR-V2X (vs.\ 60.0 ego-only) and SECOND$\to$PP reaching 79.8 AP@50 on OPV2V (vs.\ 77.8 ego-only).
BOLT thus turns the negative cooperation of unadapted fusion into a consistent positive gain across all reported pairs, establishing the feasibility of preparation-free cooperative perception.
Appendix~\ref{app:more_agents} extends this check to a three-agent OPV2V setting with two heterogeneous collaborators, and Appendix~\ref{app:fusion} repeats the PP$\to$SECOND setting with HEAL, F-Cooper, and OPV2V fusion modules.

\paragraph{Comparison with heterogeneous methods.}
Table~\ref{tab:heter} isolates the deployment question on DAIR-V2X (PP$\to$SECOND) by comparing BOLT against HEAL~\cite{HEAL} and STAMP~\cite{stamp} under both preparation-free and collaboratively trained settings.
Under the preparation-free protocol, STAMP drops to near-zero cooperative AP and HEAL retains only a degraded starting point, consistent with the fact that STAMP's adapter-reverter path was designed for a setting with cooperative preparation rather than for an unprepared interface at deployment.
From the same starting point, BOLT reaches 67.4 AP@50 through online adaptation alone, surpassing the STAMP with-base reference in this protocol-matched comparison.
This indicates that, under our deployment setting, the missing cooperative preparation stage is more critical than added model capacity.

\begin{table}[t]
\caption{Comparison with heterogeneous methods on DAIR-V2X (PP$\to$SECOND).\@
\textbf{Collab.\ Train}: requires cooperative multi-agent training.
\textbf{Extra Params}: additional parameters beyond the frozen backbone.
BOLT outperforms STAMP with base training while using no cooperative data.}\label{tab:heter}
\centering
\setlength{\tabcolsep}{5pt}
\begin{tabular}{l c c r ccc}
\toprule
\textbf{Method} & \textbf{Collab.\ Train} & \textbf{Online} & \textbf{Extra Params} & \textbf{AP@30} & \textbf{AP@50} & \textbf{AP@70} \\
\midrule
HEAL~\cite{HEAL} w/o base   & \ding{55}   & \ding{55}   & ---  & 45.9 & 44.2 & 34.4 \\
STAMP~\cite{stamp} w/o base & \ding{55}   & \ding{55}   & 0.4M & 0.0  & 0.0  & 0.0  \\
\rowcolor{gain}
\textbf{BOLT (Ours)}         & \ding{55}   & \checkmark{} & 0.9M & \textbf{72.6} & \textbf{67.4} & \textbf{53.3} \\
\cdashline{1-7}
STAMP~\cite{stamp} w/ base  & \checkmark{} & \ding{55}   & 0.4M & 60.3 & 54.5 & 32.9 \\
HEAL~\cite{HEAL} w/ base    & \checkmark{} & \ding{55}   & ---  & 82.8 & 79.1 & 64.9 \\
\bottomrule
\end{tabular}%
\end{table}

\subsection{Ablation Studies}
\begin{table}[htbp]
\caption{Ablation studies on DAIR-V2X (PP$\to$LSS-E). Default configuration highlighted in gray.}\label{tab:ablation}
\centering
\setlength{\tabcolsep}{3.5pt}
\begin{tabular}{ll r rrr r}
\toprule
\textbf{Component} & \textbf{Setting} & \textbf{Params} & \textbf{AP@30} & \textbf{AP@50} & \textbf{AP@70} & \textbf{$\Delta$@50} \\
\midrule
\multirow{3}{*}{\makecell[l]{Loss\\design}}
  & $\tau_\mathrm{hi}{=}0.0$, no boost          & —     & 60.8 & 58.6 & 48.1 & $-7.9$ \\
  & $\tau_\mathrm{hi}{=}0.3$, no boost          & —     & 67.7 & 65.2 & 54.2 & $-1.3$ \\
  & \cellcolor{gray!12}$\tau_\mathrm{hi}{=}0.3$, boost${=}0.1$\dag{} & \cellcolor{gray!12}— & \cellcolor{gray!12}\textbf{69.2} & \cellcolor{gray!12}\textbf{66.5} & \cellcolor{gray!12}\textbf{55.1} & \multicolumn{1}{c}{\cellcolor{gray!12}—} \\
\cmidrule(lr){1-7}
\multirow{5}{*}{\makecell[l]{Plugin\\size}}
  & Small ($h{=}64$, $b{=}2$)          & 0.16M  & 68.6 & 66.0 & 54.8 & $-0.5$ \\
  & \cellcolor{gray!12}Default ($h{=}128$, $b{=}3$)\dag{} & \cellcolor{gray!12}0.90M & \cellcolor{gray!12}\textbf{69.2} & \cellcolor{gray!12}\textbf{66.5} & \cellcolor{gray!12}\textbf{55.1} & \multicolumn{1}{c}{\cellcolor{gray!12}—} \\
  & Large ($h{=}256$, $b{=}4$)         & 4.76M  & 69.8 & 67.1 & 55.8 & $+0.6$ \\
  & XL ($h{=}512$, $b{=}5$)            & 23.7M  & 70.4 & 67.5 & 56.3 & $+1.0$ \\
  & XXL ($h{=}1024$, $b{=}6$)          & 113.4M & 70.9 & 68.0 & 56.9 & $+1.5$ \\
\cmidrule(lr){1-7}
\multirow{5}{*}{\makecell[l]{Plugin\\components}}
  & No plugin (direct fusion)           & —      & 58.1 & 56.2 & 47.1 & $-10.3$ \\
  & AdaIN only                          & {<}0.01M & 59.9 & 57.9 & 48.4 & $\phantom{0}-8.6$ \\
  & Adapter only (no AdaIN)             & 0.90M & 68.3 & 65.8 & 54.7 & $\phantom{0}-0.7$ \\
  & AdaIN + Adapter (no gate)           & 0.90M & 68.7 & 66.1 & 54.9 & $\phantom{0}-0.4$ \\
  & Full plugin\dag{}                  & 0.90M & \textbf{69.2} & \textbf{66.5} & \textbf{55.1} & \multicolumn{1}{c}{—} \\
\cmidrule(lr){1-7}
\multirow{5}{*}{\makecell[l]{Teacher\\confidence threshold}}
  & $\tau_\mathrm{hi}{=}0.0$                        & —     & 62.4 & 60.1 & 49.3 & $-6.4$ \\
  & $\tau_\mathrm{hi}{=}0.1$                        & —     & 69.5 & 66.7 & 55.4 & $+0.2$ \\
  & \cellcolor{gray!12}$\tau_\mathrm{hi}{=}0.3$\dag{}                 & \cellcolor{gray!12}— & \cellcolor{gray!12}\textbf{69.2} & \cellcolor{gray!12}\textbf{66.5} & \cellcolor{gray!12}\textbf{55.1} & \multicolumn{1}{c}{\cellcolor{gray!12}—} \\
  & $\tau_\mathrm{hi}{=}0.5$                        & —     & 68.9 & 66.2 & 54.9 & $-0.3$ \\
  & $\tau_\mathrm{hi}{=}0.7$                        & —     & 68.6 & 66.0 & 54.6 & $-0.5$ \\
\bottomrule
\end{tabular}
\end{table}

Table~\ref{tab:ablation} ablates the loss, plugin size, plugin components, and teacher confidence threshold on DAIR-V2X (PP$\to$LSS-E).
Confidence filtering is essential: using all teacher predictions drops AP@50 to 58.6, while the default threshold and enhancement term reach 66.5.
Most architectural gain comes from the residual adapter, with AdaIN and gating providing smaller refinements.
Larger plugins improve AP@50 only gradually despite much higher parameter counts, supporting the choice of the 0.90M default.
Because the plugin is updated online, we further report prefix AP@50 over the test stream in Appendix~\ref{app:convergence}.
Appendix~\ref{app:analyses} complements these ablations with shuffled-order robustness, a precision--recall diagnostic for the enhancement loss, and an explanation of why ego-as-teacher distillation does not cap the cooperative student at the ego detector's AP.
Appendix~\ref{app:limitations} summarizes the remaining deployment constraints, including the shared BEV grid assumption, online update cost, weak-ego cases, and the residual gap to collaboratively prepared systems.

\section{Conclusion}

We introduced preparation-free heterogeneous cooperative perception, where independently trained agents must cooperate without a shared pre-deployment protocol.
In this setting, unadapted fusion can be worse than ego-only perception because the frozen fusion module is not prepared for heterogeneous neighbor features.
BOLT addresses this failure mode by adapting only a small ego-side plugin through online ego-as-teacher distillation, while keeping the original encoders, fusion module, and detection head frozen.
Across two benchmarks and multiple encoder pairs, BOLT turns degraded preparation-free cooperation into useful cooperation and consistently surpasses ego-only performance. This suggests that lightweight online interface repair is a practical complement to prepare-first cooperative perception, particularly in open-world deployments where collaborator identities are not known in advance and offline joint training across all candidate partners is infeasible. We hope this perspective encourages further study of test-time adaptation as a first-class building block for heterogeneous multi-agent systems.

\bibliographystyle{plainnat}
\bibliography{main}

\newpage
\appendix

\section{Reproducibility and Evaluation Protocol}\label{app:impl}

\paragraph{Encoder architectures.}
We use three independently pre-trained encoders, all producing BEV feature maps of size $64 \times H \times W$:
\begin{itemize}
  \item \textbf{PointPillars (PP)}~\cite{PointPillars}: voxel size $0.4{\times}0.4{\times}5$\,m, pillar VFE with 64 filters, backbone layers $[3,5,8]$ with strides $[2,2,2]$ and filters $[64,128,256]$, multi-scale upsample to 384 channels ($128{\times}3$).
  \item \textbf{SECOND}~\cite{SECOND}: voxel size $0.1{\times}0.1{\times}0.1$\,m, mean VFE (4 point features), 3D sparse convolution ($4{\to}64$), BEV projection (128 features), backbone layers $[3]$ with stride $[1]$ and 64 filters.
  \item \textbf{LSS (camera)}~\cite{LSS}: we use both an EfficientNet backbone (LSS-E) and a ResNet-50 backbone (LSS-R50); both use image downsample factor 8, 128 image features, depth discretization via LID with 98 bins, and a BEV grid $[-102.4, 102.4]{\times}[-51.2, 51.2]$\,m at 0.4\,m resolution.
\end{itemize}

\paragraph{Preparation-free backbone construction.}
Our main preparation-free experiments do not use HEAL Stage-2 heterogeneous alignment training.
Instead, each branch is trained independently as a single-agent detector and then inserted into a frozen cooperative stack built with the same PyramidFusion architecture used by HEAL~\cite{HEAL}.
For DAIR-V2X, this means: (i) an ego PointPillars branch trained with single-agent LiDAR data, (ii) a SECOND branch trained as a general single-agent LiDAR detector, and (iii) a camera branch trained as a general single-agent detector.
All of these runs use only single-agent data ($N{=}1$), set the communication range to zero during training, and use identity aligners in the final preparation-free backbone.

\paragraph{HEAL reference baseline.}
For the collaboratively prepared HEAL reference baseline reported as an upper bound, we follow the original two-stage training protocol.
Stage~1 trains a single-modality detector (PointPillars) with the protocol alignment module using Adam optimizer (lr$=$0.002, weight decay$=$1e-4), batch size 8, for 20 epochs with multi-step LR decay (gamma$=$0.1 at epochs 10 and 15).
Stage~2 freezes the protocol module and trains additional encoder branches (SECOND or camera) to align into the same protocol space, using batch size 4, 20 epochs, and the same optimizer settings.

\paragraph{Plugin architecture.}
The adaptive plugin consists of: (1)~AdaIN normalization conditioned on ego feature statistics, with learnable per-channel blending ($C$ parameters); (2)~a residual CNN adapter with input projection $64{\to}128$, 3 residual blocks (each: $3{\times}3$ conv, GroupNorm with 16 groups, ReLU, $3{\times}3$ conv, GroupNorm, skip connection), and output projection $128{\to}64$ (zero-initialized); (3)~a per-channel gate ($C$ parameters).
Total: ${\sim}$0.9M parameters.

\paragraph{Online TTT hyperparameters.}
The adaptive plugin is optimized with Adam (lr$=$1e-4, weight decay$=$1e-4) and gradient clipping at norm 5.0.
The test stream is processed once.
Each sample is encountered once and receives one gradient update in total.
The prediction for each sample is recorded before its update, and no sample is revisited later in the stream.
The preservation loss uses confidence threshold $\tau_\mathrm{hi}{=}0.3$ with component weights: classification$=$1.0, regression$=$1.0, direction$=$0.5.
The enhancement loss uses boost weight $\lambda{=}0.1$ with confidence band $[\tau_\mathrm{lo}, \tau_\mathrm{hi}] = [0.1, 0.3]$.
The plugin's AdaIN blending coefficient is initialized at $\sigma(-10) \approx 0$ (identity-first), and the per-channel gate at $\sigma(0) = 0.5$.

\paragraph{Two-agent and multi-source evaluation.}
All main-paper tables use one ego and one active collaborator in the fusion input.
For the three-agent OPV2V experiment in Appendix~\ref{app:more_agents}, we use the multi-source extension of BOLT: each collaborator modality present in the fusion input receives its own plugin, all conditioned on the same ego feature statistics and updated under the same online protocol.

\paragraph{Detection head and loss.}
All models share the same anchor-based detection head.
Classification uses Sigmoid Focal Loss \cite{focal_loss} ($\alpha{=}0.25$, $\gamma{=}2.0$, positive weight$=$2.0).
Box regression uses Weighted Smooth-L1 Loss.
Direction classification uses Weighted Softmax Loss (2 bins, offset$=$0.7853).

\paragraph{Data splits and preprocessing.}
For DAIR-V2X, we use the official vehicle-infrastructure split.
For OPV2V, we use the default train/val/test split.
Communication range is 100\,m for all experiments.
All BEV grids use the same spatial extent per dataset.

\paragraph{Evaluation metrics.}
For both DAIR-V2X and OPV2V, evaluation follows the standard single-class vehicle-detection protocol.
Predictions are ranked by confidence score, and each predicted box is greedily matched to at most one previously unmatched ground-truth box.
A prediction is counted as a true positive only when the matched IoU exceeds the evaluation threshold; otherwise it is counted as a false positive.
IoU is computed on the BEV polygons converted from predicted and ground-truth boxes.
Precision--recall curves are accumulated over the full test split rather than averaged frame-wise, and AP is computed from the resulting curve using the VOC 2010 \cite{voc2010} interpolation protocol.
In the online setting, the prediction used for sample $t$ is always the one generated before updating the plugin on that sample.
Although some internal diagnostics may inspect distance-specific slices, all numbers reported in the main paper are overall AP values on the full evaluation set.

\section{Deployment Cost and Accounting}\label{app:efficiency}

\paragraph{Runtime efficiency.}
Table~\ref{tab:efficiency} measures the forward overhead of the plugin and the cost of one online update under the same preparation-free cooperative setting.

\begin{table}[H]
\caption{Efficiency analysis on DAIR-V2X using batch size 1 on a single NVIDIA RTX 4090.\@
\textbf{w/o} / \textbf{w/} denote forward inference without / with the plugin under the same cooperative setting.
\textbf{Online Step} measures one test-time adaptation update (teacher forward + student forward + backward + optimizer step) for the plugin-enabled model.
The plugin adds 0.90M trainable parameters in all cases.}\label{tab:efficiency}
\centering
\setlength{\tabcolsep}{4pt}
\begin{tabular}{l cc cc cc}
\toprule
\multirow{2}{*}{\textbf{Pair}} & \multicolumn{2}{c}{\textbf{Latency (ms)}} & \multicolumn{2}{c}{\textbf{FPS}} & \multirow{2}{*}{\textbf{Online Step (ms)}} & \multirow{2}{*}{\textbf{Peak Mem (GB)}} \\
\cmidrule(lr){2-3} \cmidrule(lr){4-5}
& \textbf{w/o} & \textbf{w/} & \textbf{w/o} & \textbf{w/} & & \\
\midrule
PP$\to$SECOND  & 69.4  & 85.8  & 14.4 & 11.7 & 172.1 & 2.09 \\
PP$\to$LSS-E   & 144.7 & 155.8 & 6.9  & 6.4  & 230.7 & 2.14 \\
PP$\to$LSS-R50 & 128.3 & 149.1 & 7.8  & 6.7  & 224.6 & 2.09 \\
\bottomrule
\end{tabular}
\end{table}

The plugin adds only modest forward overhead relative to frozen preparation-free inference, increasing latency by 11.1--20.8\,ms across the three representative DAIR-V2X pairs while keeping throughput at 6.4--11.7\,FPS.\@
The main cost comes from the online update itself, which takes 172.1--230.7\,ms with about 2.1\,GB peak memory.
In other words, BOLT pays for deployment-time adaptation primarily in optimization time, not in a large parameter footprint.

\paragraph{Cost accounting protocol.}
We do not count the pretraining cost of the single-agent detectors themselves.
That cost is shared by all methods in our deployment setting, where each agent is assumed to already carry an independently trained perception backbone before cooperation begins.
Instead, we count only the additional method-specific cost required to make cooperation work.
This includes extra offline alignment or fusion training, any dependence on cooperative multi-agent training data, and any online deployment-time optimization.
Under this protocol, BOLT adds no offline cooperative training cost but does incur an online adaptation cost, reported in Table~\ref{tab:efficiency}.
Table~\ref{tab:cost_accounting} summarizes the accounting rule we follow throughout the comparisons.

\begin{table}[H]
\caption{Cost accounting protocol used in our comparisons.}\label{tab:cost_accounting}
\centering
\small
\resizebox{\columnwidth}{!}{%
\begin{tabular}{p{0.30\columnwidth} c p{0.52\columnwidth}}
\toprule
\textbf{Cost Item} & \textbf{Counted?} & \textbf{Reason} \\
\midrule
Single-agent detector pretraining & No & Common prerequisite for all deployed agents; not specific to enabling cooperation. \\
Additional offline alignment / fusion training & Yes & Method-specific extra cost required before heterogeneous cooperation can work. \\
Cooperative multi-agent training data & Yes & Additional supervision requirement beyond independently trained single-agent models. \\
Online test-time adaptation & Yes & Deployment-time cost directly incurred by the method. \\
\bottomrule
\end{tabular}
}
\end{table}

\paragraph{Approximate wall-clock references.}
To complement the accounting view above, Table~\ref{tab:wallclock_cost} gives rough wall-clock references on DAIR-V2X (PP$\to$SECOND) using one NVIDIA RTX 4090 under our implementation.
These values are intended as order-of-magnitude references rather than a strict benchmark.
The shared single-agent detector pretraining is listed separately because it is a common prerequisite and is excluded from the main method-specific accounting.

\begin{table}[H]
\caption{Approximate wall-clock cost references on DAIR-V2X (PP$\to$SECOND) using one NVIDIA RTX 4090. Times are rough implementation-level estimates intended for scale comparison only, not a strict benchmark.}\label{tab:wallclock_cost}
\centering
\small
\setlength{\tabcolsep}{4pt}
\resizebox{\columnwidth}{!}{%
\begin{tabular}{p{0.18\columnwidth} p{0.34\columnwidth} c p{0.26\columnwidth}}
\toprule
\textbf{Method} & \textbf{Additional Cooperative Cost} & \textbf{Approx. Time} & \textbf{Interpretation} \\
\midrule
Shared single-agent detector & Shared prerequisite only (excluded from main cost accounting) & ${\sim}$1.5 h & Listed for context only. \\
HEAL & Homogeneous base + heterogeneous stage-2 alignment & ${\sim}$12 h & Two offline stages dominate the cost. \\
STAMP & Offline cooperative training & ${\sim}$8--10 h & One main cooperative training stage. \\
PHCP & Cooperative base + label-free adaptation & ${\sim}$8--10 h + ${\sim}$10--30 min & Short adaptation stage, but still needs a cooperative base and unlabeled collaborator data. \\
\textbf{BOLT (ours)} & Online TTT only & ${\sim}$30 min & No offline cooperative training beyond the shared single-agent detectors. \\
\bottomrule
\end{tabular}
}
\end{table}

\section{Main-Result Extensions}

\subsection{Beyond Two Agents}\label{app:more_agents}

To test whether the same preparation-free interface problem persists beyond the two-agent setting, we additionally evaluate a three-agent heterogeneous configuration on OPV2V.
The ego vehicle uses PointPillars, while the two collaborators use LSS-EfficientNet and SECOND, respectively.
We use the multi-source extension of BOLT, with one dedicated plugin for each collaborator modality.
Because scenes with four or five active vehicles are rare in the OPV2V test split, we report only the three-agent case.
Table~\ref{tab:three_agent} shows the same qualitative pattern as in the main paper: unadapted preparation-free cooperation is unreliable, whereas the plugin restores strong positive cooperation even when two heterogeneous neighbors are present.

\begin{table}[H]
\caption{Three-agent heterogeneous collaboration on OPV2V using the multi-source extension of BOLT. The ego uses PointPillars, and the two collaborators use LSS-EfficientNet and SECOND. Because scenes with four or five active vehicles are rare in the OPV2V test split, we report only the three-agent case.}\label{tab:three_agent}
\centering
\small
\setlength{\tabcolsep}{4.5pt}
\begin{tabular}{l ccc ccc c}
\toprule
\multirow{2}{*}{\textbf{Setting}} & \multicolumn{3}{c}{\textbf{w/o Plugin}} & \multicolumn{3}{c}{\textbf{w/ Plugin}} & \multirow{2}{*}{\textbf{$\Delta$AP@50}} \\
\cmidrule(lr){2-4} \cmidrule(lr){5-7}
& AP@30 & AP@50 & AP@70 & AP@30 & AP@50 & AP@70 & \\
\midrule
PP (ego) + LSS-E + SECOND & 43.5 & 43.0 & 35.0 & \cellcolor{gain}\textbf{82.2} & \cellcolor{gain}\textbf{80.5} & \cellcolor{gain}\textbf{65.3} & \g{+37.5} \\
\bottomrule
\end{tabular}
\end{table}

\subsection{Fusion-Backbone Compatibility}\label{app:fusion}

\begin{table}[H]
\caption{Plugin performance across three fusion backbones on DAIR-V2X (PP$\to$SECOND). The plugin improves all three.}\label{tab:fusion}
\centering
\begin{tabular}{l ccc ccc c}
\toprule
\multirow{2}{*}{\textbf{Fusion}} & \multicolumn{3}{c}{\textbf{w/o Plugin}} & \multicolumn{3}{c}{\textbf{w/ Plugin}} & \multirow{2}{*}{\textbf{$\Delta$AP@50}} \\
\cmidrule(lr){2-4} \cmidrule(lr){5-7}
& AP@30 & AP@50 & AP@70 & AP@30 & AP@50 & AP@70 & \\
\midrule
HEAL~\cite{HEAL} & 46.0 & 44.2 & 34.4 & \cellcolor{gain}\textbf{72.6} & \cellcolor{gain}\textbf{67.4} & \cellcolor{gain}\textbf{53.3} & \g{+23.2} \\
F-Cooper~\cite{Fcooper}   & 21.2 & 20.3 & 16.8 & \cellcolor{gain}\textbf{70.5} & \cellcolor{gain}\textbf{66.5} & \cellcolor{gain}\textbf{54.4} & \g{+46.2} \\
OPV2V~\cite{OPV2V} & 69.2 & 66.2 & 56.7 & \cellcolor{gain}\textbf{71.6} & \cellcolor{gain}\textbf{67.9} & \cellcolor{gain}\textbf{58.0} & \g{+1.7} \\
\bottomrule
\end{tabular}%
\end{table}

Table~\ref{tab:fusion} shows that the plugin is not tied to a single fusion backbone.
F-Cooper is the most sensitive to preparation-free mismatch, dropping to 20.3 AP@50 without adaptation, and it gains the most once the interface is repaired.
OPV2V provides the strongest no-plugin baseline at 66.2 AP@50, yet the plugin still improves it to 67.9.
HEAL lies between these two cases.
Overall, BOLT behaves as a general interface adapter between the neighbor encoder and the fusion module rather than as a backbone-specific redesign.

\subsection{Extended Qualitative Results}\label{app:qual}

\begin{figure}[H]
  \centering
  \includegraphics[width=\textwidth]{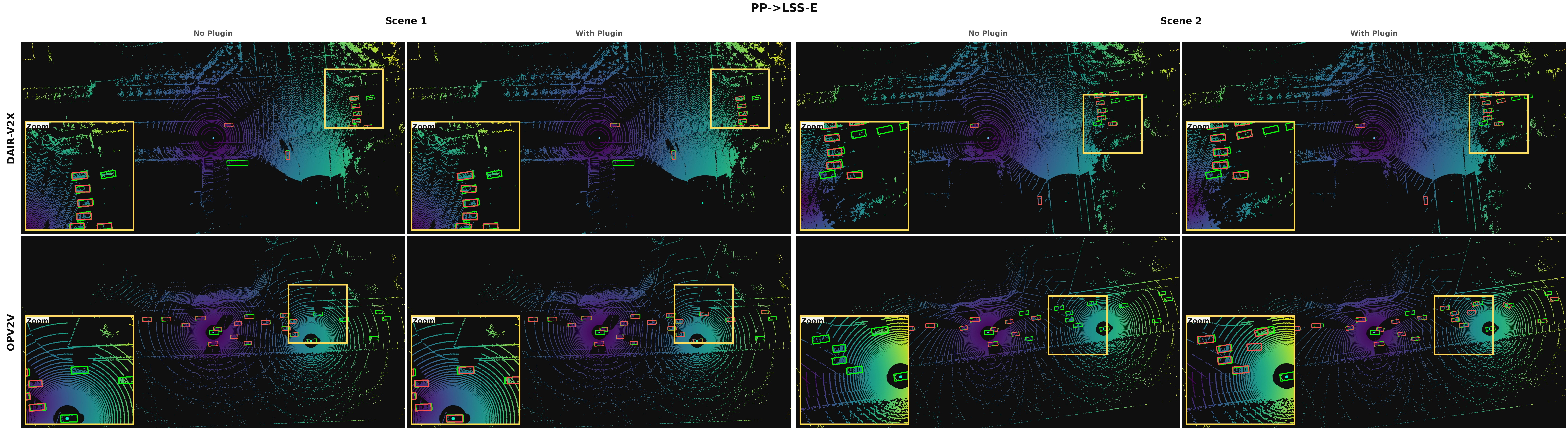}
  \caption{Additional qualitative BEV results for PP$\to$LSS-E on DAIR-V2X (top) and OPV2V (bottom). Each panel shows the full BEV view together with a zoomed ROI from the highlighted region.}\label{fig:bev_pp_camera_e}
\end{figure}

\begin{figure}[H]
  \centering
  \includegraphics[width=\textwidth]{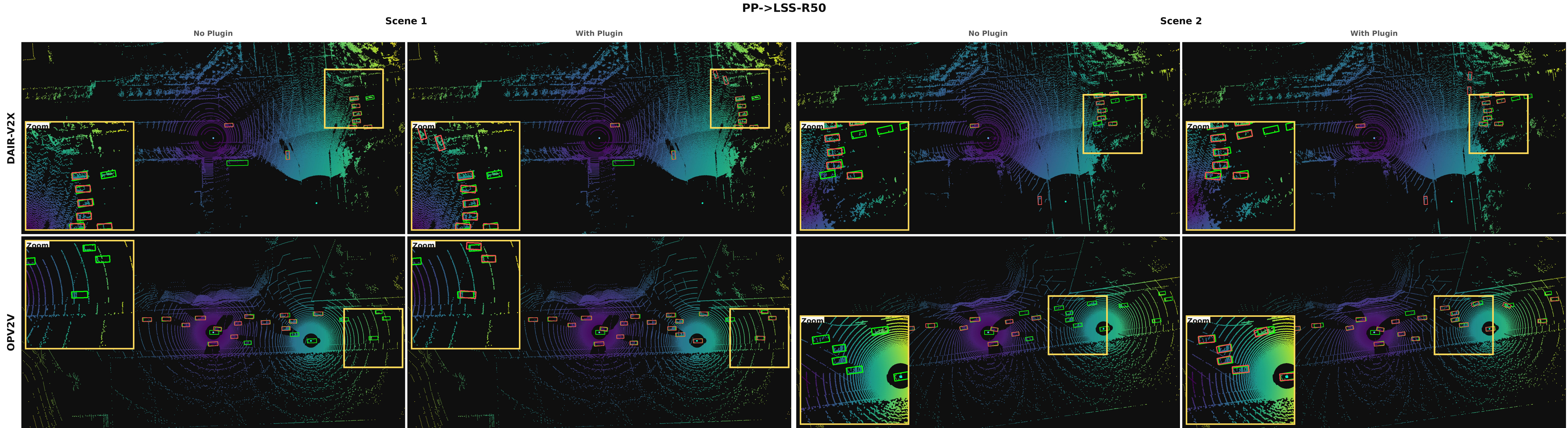}
  \caption{Additional qualitative BEV results for PP$\to$LSS-R50 on DAIR-V2X (top) and OPV2V (bottom). Each panel shows the full BEV view together with a zoomed ROI from the highlighted region.}\label{fig:bev_pp_camera_r50}
\end{figure}

\begin{figure}[H]
  \centering
  \includegraphics[width=\textwidth]{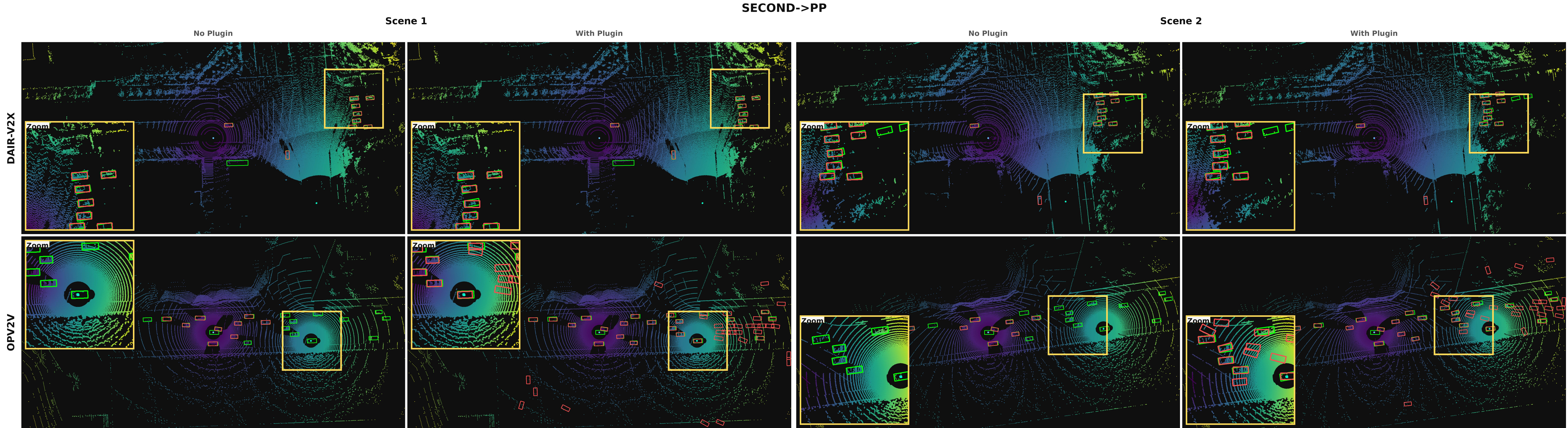}
  \caption{Additional qualitative BEV results for SECOND$\to$PP on DAIR-V2X (top) and OPV2V (bottom). Each panel shows the full BEV view together with a zoomed ROI from the highlighted region.}\label{fig:bev_second_pp}
\end{figure}

\begin{figure}[H]
  \centering
  \includegraphics[width=\textwidth]{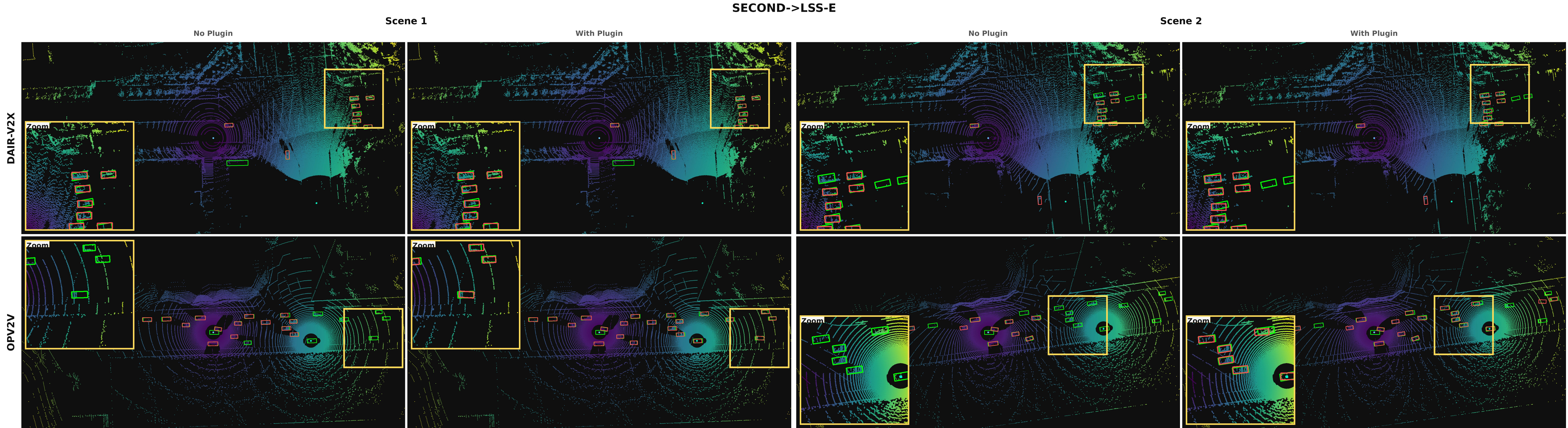}
  \caption{Additional qualitative BEV results for SECOND$\to$LSS-E on DAIR-V2X (top) and OPV2V (bottom). Each panel shows the full BEV view together with a zoomed ROI from the highlighted region.}\label{fig:bev_second_camera_e}
\end{figure}

The appendix figures provide additional qualitative examples from Table~\ref{tab:main} that complement the PP$\to$SECOND visualization in Figure~\ref{fig:bev} of the main paper. Figures~\ref{fig:bev_pp_camera_e}--\ref{fig:bev_second_camera_e} cover four additional pairs: PP$\to$LSS-E, PP$\to$LSS-R50, SECOND$\to$PP, and SECOND$\to$LSS-E. Each figure keeps the same layout: DAIR-V2X on the first row, OPV2V on the second row, and for each scene a full BEV view together with a zoomed ROI from the highlighted region.

\section{Online Adaptation and Design Diagnostics}\label{app:convergence}

\begin{figure}[H]
  \centering
  \includegraphics[width=\textwidth]{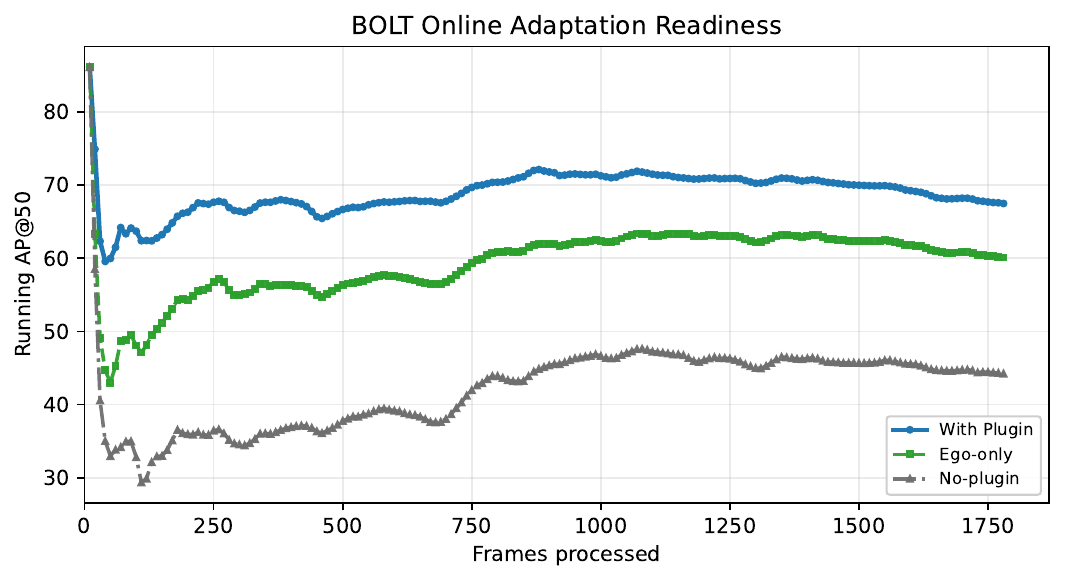}
  \caption{Dynamic online convergence of BOLT on DAIR-V2X (PP$\to$SECOND). AP@50 is computed on the prefix processed so far and plotted every 10 samples for BOLT, a standalone PP ego-only detector, and no-plugin under the same stream order. BOLT quickly recovers from degraded preparation-free cooperation, overtakes the standalone ego-only baseline, and remains stable thereafter.}\label{fig:convergence}
\end{figure}

Figure~\ref{fig:convergence} shows that adaptation stabilizes quickly on the online DAIR-V2X stream for PP$\to$SECOND.\@
Each point is a prefix-matched AP@50 value computed on all samples processed so far under the same stream order.
We plot three curves, BOLT, a standalone PP ego-only detector, and no-plugin, every 10 samples.
BOLT rises rapidly from the degraded no-plugin baseline, overtakes the standalone ego-only reference after the early transient, and then remains on a stable plateau with only mild late-stream fluctuation, indicating stable adaptation rather than catastrophic drift.

\subsection{Ordering and Loss Diagnostics}\label{app:analyses}

\paragraph{Sensitivity to sample ordering.}
Because BOLT performs online adaptation sequentially, the order in which test samples are presented may affect the final AP.\@
We therefore treat ordering robustness as an additional diagnostic rather than a primary benchmark.
Since the main-paper results are reported under the default dataset order, Table~\ref{tab:ordering} shows both that default-order result and a shuffled-order stress test on DAIR-V2X (PP$\to$SECOND).
The default order, which is the one used in the main tables, reaches 67.4 AP@50.
Across five shuffled test-stream orders, the mean becomes $67.3 \pm 0.2$, and the AP@50 range remains narrow at [67.0, 67.4].
This indicates that stream order does matter in online adaptation, while the variation across shuffled orders is itself small once the order is fixed.

\begin{table}[H]
\caption{Sensitivity to test-stream ordering on DAIR-V2X (PP$\to$SECOND). The main-paper result uses the default dataset order; shuffled orders are reported as an additional robustness diagnostic.}\label{tab:ordering}
\centering
\small
\setlength{\tabcolsep}{4.5pt}
\begin{tabular}{l ccc}
\toprule
\textbf{Stream Order} & \textbf{AP@30} & \textbf{AP@50} & \textbf{AP@70} \\
\midrule
Default order (main tables) & 72.6 & 67.4 & 53.3 \\
5 shuffled orders (mean$\pm$std) & 71.7$\pm$0.2 & 67.3$\pm$0.2 & 51.4$\pm$0.1 \\
Shuffled range & [71.4, 71.9] & [67.0, 67.4] & [51.2, 51.5] \\
\bottomrule
\end{tabular}
\end{table}

\begin{figure}[htbp]
  \centering
  \includegraphics[width=0.82\textwidth]{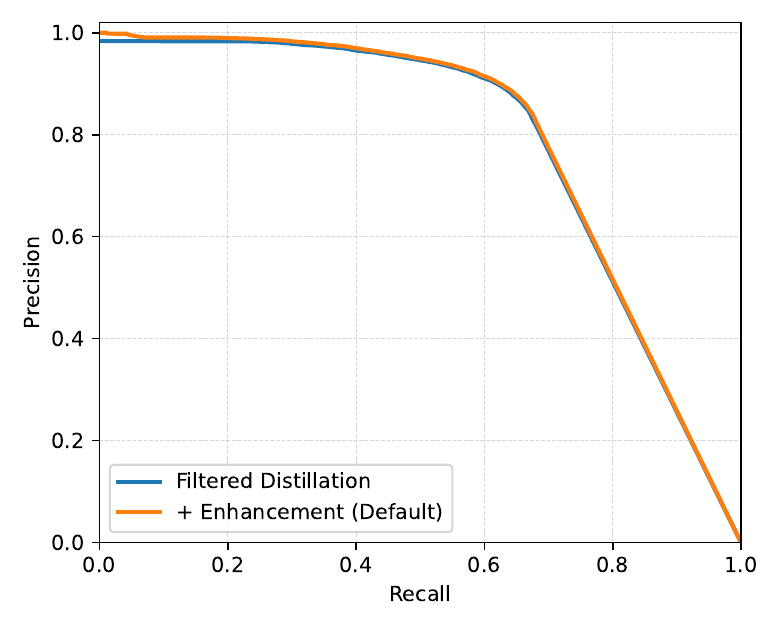}
  \caption{Precision--recall curves at IoU 0.5 on DAIR-V2X (PP$\to$LSS-E). Adding the enhancement loss ($\tau_\mathrm{hi}{=}0.3$, boost${=}0.1$) improves the high-confidence part of the curve and slightly increases AP@50, while the low-precision tail remains close to the no-boost baseline.}\label{fig:pr_analysis}
\end{figure}

\paragraph{Precision--recall behavior of enhancement loss.}
The enhancement loss (Eq.~\ref{eq:enhance}) raises classification confidence in uncertain regions, so a natural concern is whether it does so by creating extra false positives.
Figure~\ref{fig:pr_analysis} compares filtered distillation ($\tau_\mathrm{hi}{=}0.3$, no boost) with the default loss ($\tau_\mathrm{hi}{=}0.3$, boost${=}0.1$) on DAIR-V2X (PP$\to$LSS-E) at IoU 0.5.
The default loss improves the high-confidence part of the curve and slightly raises AP@50, while the low-precision tail remains close to the no-boost baseline.
This indicates better ranking of confident detections rather than a large shift in the recall ceiling.

\paragraph{Why ego-as-teacher does not cap student performance.}
A natural concern about ego-as-teacher distillation is whether the cooperative student is implicitly capped at the ego detector's accuracy.
It is not, because the two losses act on disjoint spatial regions.
The preservation loss (Eq.~\ref{eq:preserve}) is gated by the ego confidence threshold $\sigma(\mathbf{t}_\mathrm{cls}^i) > \tau_\mathrm{hi}$ and only locks anchors where the ego is already reliable; at all other locations the student is unconstrained by the teacher.
The enhancement loss (Eq.~\ref{eq:enhance}) operates exactly on this complementary region, encouraging the student to upgrade low-confidence ego responses whenever the neighbor feature provides corroborating evidence.
Objects that are occluded or distant from the ego viewpoint typically yield weak ego responses but strong neighbor features, and the fused representation can therefore recover detections that the ego alone would miss or score too weakly.
The empirical confirmation is in the main results: every evaluated pair surpasses the ego-only baseline (Table~\ref{tab:main}).

\section{Extended Limitations}\label{app:limitations}

\paragraph{Shared BEV spatial resolution.}
BOLT assumes that the ego feature and the received neighbor BEV feature share the same spatial resolution $C{\times}H{\times}W$ after their respective encoders, so that the plugin's AdaIN, residual adapter, and per-channel gate can operate element-wise on each received neighbor feature before fusion.
In practice, most cooperative perception pipelines built on OpenCOOD already enforce this resolution through a shared BEV grid configuration, which is why the assumption holds across all encoder pairs we evaluate.
Agents whose single-agent detectors use genuinely different BEV grids (different voxel sizes or feature-map strides) would require an additional spatial alignment step, for example a resize or a lightweight spatial resampler, before the plugin can be applied.
This is a limitation of the current plugin design rather than of the preparation-free setting itself.

\paragraph{Extra inference cost from online updates.}
Online adaptation is not free.
Table~\ref{tab:efficiency} shows that the plugin adds 11.1--20.8\,ms of forward overhead relative to frozen preparation-free inference, and the online update step itself takes 172.1--230.7\,ms with about 2.1\,GB peak memory on a single RTX 4090.
In deployment scenarios with tight real-time budgets, this cost may be partially amortized by updating only at a lower rate (e.g., every $k$ samples) rather than every frame, at the expense of slightly slower convergence.
We view this latency--convergence trade-off as a deployment tuning knob rather than a fundamental blocker.

\paragraph{Weak ego encoder.}
BOLT relies on the ego detector as supervision.
When the ego encoder itself is weak, most notably in Camera$\to$LiDAR settings, the teacher signal is weak as well and the achievable cooperation gain shrinks.
In these cases, ego-only AP is already much lower than in LiDAR$\to$X settings, and online adaptation yields smaller absolute gains.
This is a direct limitation of the ego-as-teacher design: the student cannot reliably learn detections the teacher does not support.
We therefore treat Camera$\to$LiDAR as a limitation case rather than a core benchmark.

\paragraph{Residual gap to the homogeneous cooperative upper bound.}
Although BOLT consistently surpasses the ego-only baseline across all evaluated pairs, a gap to the homogeneous cooperative upper bound (e.g., HEAL with cooperative pre-training in Table~\ref{tab:heter}) remains.
We attribute this gap to three deliberate design choices of the preparation-free setting itself.
First, the plugin is intentionally lightweight ($\sim$0.9M parameters) so that adaptation can converge from a single online pass; a larger plugin closes part of the gap (Table~\ref{tab:ablation}, Plugin size) but at higher latency and memory cost.
Second, the teacher is bounded by the ego detector and provides no supervision for objects the ego cannot see at all, so detections that are exclusively visible to the neighbor cannot be recovered through distillation.
Third, the fusion module is held frozen at its single-agent configuration, so any residual protocol mismatch internal to the fusion module itself is not corrected by the plugin.
Closing this gap further, for example by jointly adapting a small portion of the fusion module or by introducing a lightweight cross-agent consistency signal, is a natural next step rather than a prerequisite for the validity of the preparation-free setting.

\end{document}